# An Analysis on Ensemble Learning optimized Medical Image Classification with Deep Convolutional Neural Networks

*Dominik Müller[1,2], Iñaki Soto-Rey[1,2] and Frank Kramer[1]*

[1] IT-Infrastructure for Translational Medical Research, University of Augsburg, Germany
[2] Medical Data Integration Center, Institute for Digital Medicine, University Hospital Augsburg, Germany

## Abstract

**Background:** Novel and high-performance medical image classification pipelines are heavily utilizing ensemble learning strategies. The idea of ensemble learning is to assemble diverse models or multiple predictions and, thus, boost prediction performance. However, it is still an open question to what extent as well as which ensemble learning strategies are beneficial in deep learning based medical image classification pipelines.
**Methods:** In this work, we proposed a reproducible medical image classification pipeline for analyzing the performance impact of the following ensemble learning techniques: Augmenting, Stacking, and Bagging. The pipeline consists of state-of-the-art preprocessing and image augmentation methods as well as 9 deep convolution neural network architectures. It was applied on four popular medical imaging datasets with varying complexity. Furthermore, 12 pooling functions for combining multiple predictions were analyzed, ranging from simple statistical functions like unweighted averaging up to more complex learning-based functions like support vector machines.
**Results:** Our results revealed that Stacking achieved the largest performance gain of up to 13% F1-score increase. Augmenting showed consistent improvement capabilities by up to 4% and is also applicable to single model based pipelines. Cross-validation based Bagging demonstrated significant performance gain close to Stacking, which resulted in an F1-score increase up to +11%. Furthermore, we demonstrated that simple statistical pooling functions are equal or often even better than more complex pooling functions.
**Conclusions:** We concluded that the integration of ensemble learning techniques is a powerful method for any medical image classification pipeline to improve robustness and boost performance.

## 1. Introduction

The field of automated medical image analysis has seen rapid growth in recent years [1–3]. The utilization of deep neural networks became one of the most popular and widely applied algorithms for computer vision tasks [2]. A starting point for this trend relies on deep convolutional neural network architectures. These architectures demonstrated powerful prediction capabilities and achieved similar performance as clinicians [2,4]. The integration of deep learning based automated medical image analysis in the clinical routine is currently a highly popular research topic. The subfield medical image classification (MIC) aims to label a complete image to predefined classes, e.g. to a diagnosis or a condition. The idea is to use these models as clinical decision support for clinicians in order to improve diagnosis reliability or automate time-consuming processes [2,5].

Recent studies showed that the most successful and accurate MIC pipelines are also heavily based on ensemble learning strategies [6–13]. In the machine learning field, the aim is to find a suitable hypothesis that maximizes prediction correctness. However, finding the optimal hypothesis is difficult which is why the strategy was evolved to combine multiple hypotheses into a superior predictor closer to an optimal hypothesis. In the context of deep convolutional neural networks, hypotheses are represented through fitted neural network models. Thus, ensemble learning is defined as the combination of models to yield better prediction performance. The integration of ensemble learning strategies in a deep learning based pipeline is called deep ensemble learning. Various recent studies successfully utilized this strategy to improve the performance and robustness of their MIC pipeline [6–17]. The underlying techniques of these deep ensemble learning based pipelines are ranging from the combination of different model types like in the studies Rajaraman et al. [18] and Pham et al. [19] to inference improvement of a single model like in Galdran et al. [20]. Furthermore, medical imaging datasets are commonly quite small, which is why ensemble learning techniques for efficient training data usage are especially popular as demonstrated in Bibaut et al. [14] and Müller et al. [21].

Empirically, ensemble learning based pipelines tend to be superior according to the assumption that the assembling of diverse models has the advantage to combine their strengths in focusing on different features whereas balancing out the individual incapability of a model [14,22–24]. However, it is still an open question to

Correspondence: dominik.mueller@informatik.uni-augsburg.de

what extent as well as which ensemble learning strategies are beneficial in deep learning based MIC pipelines. Even so, the field and idea of general ensemble learning is not novel, the impact of ensemble learning strategies in deep learning based classification has not been adequately analyzed in the literature, yet. Whereas multiple authors provide extensive reviews on general ensemble learning like Ganaiea et al. [24], only a handful of works started to survey the deep ensemble learning field. While Cao et al. reviewed deep learning based ensemble learning methods specifically in bioinformatics [25], Sagi et al. [26], Ju et al. [14], and Kandel et al. [27] started to provide descriptions or analysis on general deep ensemble learning methods.

In this study, we push towards setup a reproducible analysis pipeline to reveal the impact of ensemble learning techniques on medical image classification performance with deep convolution neural networks. By computing the performance of multiple ensemble learning techniques, we want to compare them to a baseline pipeline and, thus, identify possible performance gain. Furthermore, we explore the possible performance impact on multiple medical datasets from diverse modalities ranging from histology to X-ray imaging. Our experiments aim to help understand the beneficial as well as unfavorable influences of different ensemble learning techniques on model performance. This study contributes to the field of deep ensemble learning and provides the missing overview of state-of-the-art ensemble learning techniques for deep learning based MIC.

Our manuscript is organized as follows: Section 1 introduces medical image classification, the field of ensemble learning and our research question. In Section 2, we describe our proposed pipeline including the datasets, preprocessing methods, deep convolutional neural network architectures, ensemble learning strategies, and pooling functions. In Section 3, we report the experimental results and discuss these in detail in Section 4. In Section 5, we conclude our paper and give insights on future work. The Appendix contains further information on the availability of our trained models, all result data and the code used in this research.

## 2. Methods and Materials

### 2.1 Datasets

For increased result reliability and robustness, we analyzed multiple public MIC datasets. The datasets differ in sample size, modality, feature type of interest and noisiness. An overview of all datasets can be seen in Table 1, as well as exemplary samples in Figure 1.

2.1.1 CHMNIST: The image analysis of histological slides is an essential part in the field of pathology. The CHMNIST dataset consists of image patches generated from histology slides of patients with colorectal cancer [28,29]. These patches were annotated in eight distinct classes: Tumor epithelium, simple stroma (homogeneous composition), complex stroma (containing single tumor cells and/or immune cells), immune cells, debris (including necrosis, hemorrhage and mucus), normal mucosal glands, adipose tissue and background (no tissue) [28,29]. The dataset contains in total 5,000 images in Red-Green-Blue (RGB) color encoding with 625 images for each class and a unified resolution of 150x150 pixels. The slides were generated via an Aperio ScanScope microscope with a 20x magnification from the pathology archive of University Medical Center Mannheim and Heidelberg University [28].

2.1.2 COVID: X-ray imaging is one of the key modalities in the field of medical image analysis and is crucial in modern healthcare. Furthermore, X-ray imaging is a widely favored alternative to reverse transcription polymerase chain reaction testing for the coronavirus disease (COVID-19) [30,31]. Researchers from Qatar, Doha, Dhaka, Bangladesh, Pakistan and Malaysia have created a dataset of thorax X-ray images for COVID-19 positives cases along with healthy control and other viral pneumonia cases [30]. The X-ray scans were gathered and annotated from 6 different radiographic databases or sources like the Italian Society of Medical and Interventional Radiology (SIRM) COVID-19 Database [30,32]. The dataset consists of in total 2,905 grayscale images with 219 COVID-19 positive, 1,345 viral pneumonia and 1,341 control cases.

2.1.3 ISIC: Melanoma, appearing as pigmented lesions on the skin, is a major public health problem with more than new 300,000 cases per year and is responsible for the majority of skin cancer deaths [33]. Dermoscopy is the field of early melanoma detection, which can be either performed manually by expert visual inspection or automatically by MIC via high-resolution cameras. The International Skin Imaging Collaboration (ISIC) hosts the largest publicly available collection of quality-controlled images of skin lesions [33]. The 2019 release of their archive consists of in total 25,331 RGB images which were classified in the following 8 classes: Melanoma (MEL), melanocytic nevus (NV), basal cell carcinoma (BCC), actinic keratosis (AK), benign keratosis (BKL), dermatofibroma (DF), vascular lesion (VASC) and squamous cell carcinoma (SCC) [34–36].

2.1.4 DRD: Diabetic retinopathy is the leading cause of blindness and is estimated to affect over 93 million people worldwide[37]. The detection of diabetic retinopathy is mostly done via a time-consuming manual inspection by a clinician or ophthalmologist with the help of a fundus camera[21]. In order to contribute to research for automated diabetic retinopathy detection (DRD) algorithms, the California Healthcare Foundation and EyePACS created a public dataset consisting of 35,126 RGB fundus images [37,38]. These were annotated in the following five classes according to disease severity: No DR, Mild, Moderate, Severe, Proliferative DR. It has to be noted that the authors pointed out the real-world aspect of this dataset which includes various types of noise like artifacts, out of focus, under-/overexposed images and incorrect annotations [37].

**Table 1:** Overview of utilized datasets with descriptive details and sampling distributions. The noisiness of a dataset is a subjective impression based on the best achieved performance in the literature for this dataset.

| Dataset | Modality | Noisiness | Classes | Number of Samples | | | |
|---|---|---|---|---|---|---|---|
| | | | | model-train | model-val | ensemble-train | testing |
| **CHMNIST** | Histology | Small | 8 | 3,250 | 501 | 500 | 749 |
| **COVID** | X-ray | Small | 3 | 1,889 | 291 | 290 | 435 |
| **ISIC** | Dermoscopy | Medium | 8 | 16,466 | 2,533 | 2,533 | 3,799 |
| **DRD** | Ophthalmoscopy | Strong | 5 | 22,832 | 3,513 | 3,513 | 5,268 |

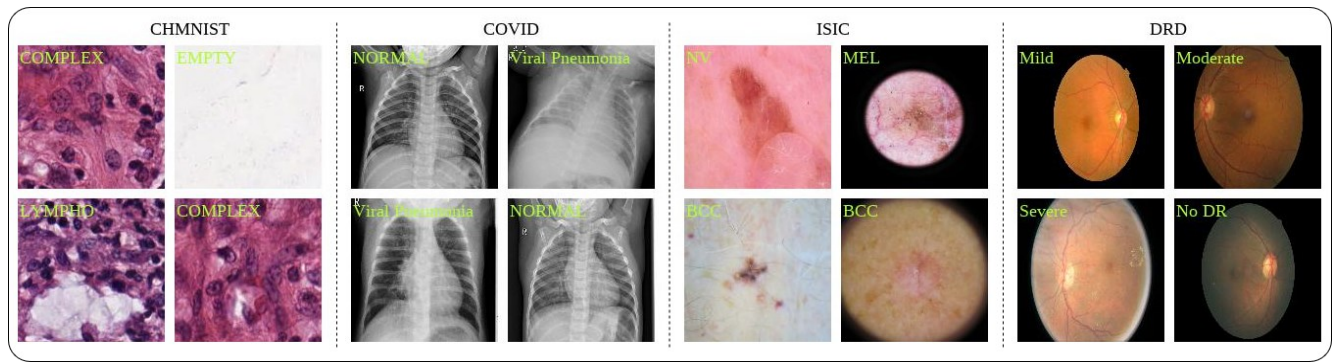

**Figure 1:** Exemplary sections of the four used datasets used in this analysis: CHMNIST (histology), COVID (X-ray), ISIC (dermoscopy) and DRD (ophthalmoscopy).

## 2.2 Sampling and Preprocessing

In order to ensure a reliable evaluation of our models, we sampled each dataset with the following distribution strategy: For model training, 65% of each dataset was used (called 'model-train') whereas 10% of all samples were used as a validation set during the training process (called 'model-val') to allow validation monitoring for callback strategies. The only exception for this 'model-train' and 'model-val' sampling strategy occurred in the Bagging experiment, in which the two sets were combined and sampled according to a 5-fold cross-validation (75% in total of a dataset with 60% as training and 15% as validation for each fold). For possible training of ensemble learning pooling methods, another 10% of a total dataset was reserved (called 'ensemble-train'). For the final in detail evaluation on a separate hold-out set, the remaining 15% of each dataset was sampled as testing set (called 'testing').

We applied the following preprocessing methods for enhancement of the pattern-finding process of our deep learning models as well as to increase data variability. Our pipeline utilized extensive real-time (also called online-) image augmentation during the training phase to allow the model seeing novel and unique images in each epoch. The augmentation was performed with Albumentations [39] and consisted of the following techniques: Flipping, rotations as well as alterations in brightness, contrast, saturation and hue. Furthermore, all images were squared padded for avoiding aspect ratio loss. In the posterior resizing, the image resolutions were reduced to the model architecture default input sizes, which were commonly 224x224 pixels except for EfficientNetB4 with 380x380, as well as InceptionResNetV2 and Xception with 299x299 pixels [40–42]. Before passing the images into the model, we applied value intensity normalization. The intensities were zero-centered via Z-Score normalization based on the mean and standard deviation computed on the ImageNet dataset [43].

## 2.3 Deep Learning Models

For computer vision tasks like image classification, deep convolutional neural networks are state-of-the-art and unmatched in accuracy and robustness [5,44–46]. Rather than focusing on a single model architecture for our analysis, we trained diverse classification architectures to ensure result reliability. The following architecture were selected: DenseNet121 [47], EfficientNetB4 [41], InceptionResNetV2 [42], MobileNetV2 [48], ResNeXt101 [49], ResNet101 [50], VGG16 [51], Xception [40] and a custom Vanilla architecture for comparison. The Vanilla architecture consisted of 4 convolutional layers with each followed by a max-pooling layer. The utilized classification head for all architectures applied a global average pooling, a dense layer, a dropout layer, another dense layer and a softmax activation layer for the final class probabilities. The selected architectures represent the large diversity of popular and widely applied types of deep learning models for image classification. These strongly vary in the number of model parameters as well as neural network layers, input sizes, underlying composition techniques as well as functionality principles, and overall complexity. This allows a clearer analysis of the ensemble learning impact without architecture-related biases. Further details on the architectures and their differences can be found in the excellent reviews of Bressem et al. [19] and Alzubaidi et al. [52]. For implementation, we used our in-house developed framework AUCMEDI which is built on TensorFlow [53].

We utilized a transfer learning strategy by pretraining all models on the ImageNet dataset [43]. For the fitting process, the architecture layers were frozen at first except for the classification head and unfrozen, again, for fine-tuning. Whereas the frozen transfer learning phase was performed for 10 epochs using the Adam optimization with an initial learning rate of 1-E04, the fine-tuning phase stopped after a maximal training time of 1000 epochs (including the 10 epochs for transfer learning). The fine-tuning phase also utilized a dynamic learning rate for the Adam optimization [54] starting from 1-E05 to a maximum decrease to 1-E07 by a decreasing factor of 0.1 after 8 epochs without improvement on the monitored validation loss. As loss function for model training, we used the weighted Focal loss from Lin et al. [55].

$$\text{FL}(p_t) = -\alpha_t(1-p_t)^\gamma \log(p_t) \quad (1)$$

In the above formula, $p_t$ is the probability for the correct ground truth class *t*, *γ* a tunable focusing parameter (which we set to 2.0) and $\alpha_t$ the associated weight for class *t* [55]. The class weights were computed based on the class distribution in the corresponding 'model-train' sampling set. Furthermore, an early stopping and model checkpoint technique was applied for the fine-tuning phases, stopping after 15 epochs without improvement and saving the best model measured based on validation loss monitoring. The complete analysis was performed with a batch size of 28 and run parallelized on a workstation with 4x NVIDIA Titan RTX with each 24GB VRAM, Intel(R) Xeon(R) Gold 5220R CPU @ 2.20GHz with 96 cores and 384GB RAM.

## 2.4 Ensemble Learning Techniques

As stated in the Introduction, deep ensemble learning is traditionally defined as building an ensemble of multiple predictions originating from different deep convolutional neural network models [24]. However, recent novel techniques necessitate redefining ensemble learning in the deep learning context as combining information, most commonly predictions, for a single inference. This information or predictions can either originate from multiple distinct models or just a single model. In this analysis, we explored the performance impact of the ensemble learning techniques: Augmenting, Bagging, and Stacking. We excluded the Boosting technique, which is also commonly used in general ensemble learning. The reason for this is that Boosting is not feasibly applicable for image classification with deep convolutional neural networks due to the extreme increase in training time [24,26]. An overview diagram of the four techniques can be seen in Figure 2. For comparison, we setup Baseline models for all architectures to identify possible performance gain or loss tendencies through the ensemble learning techniques.

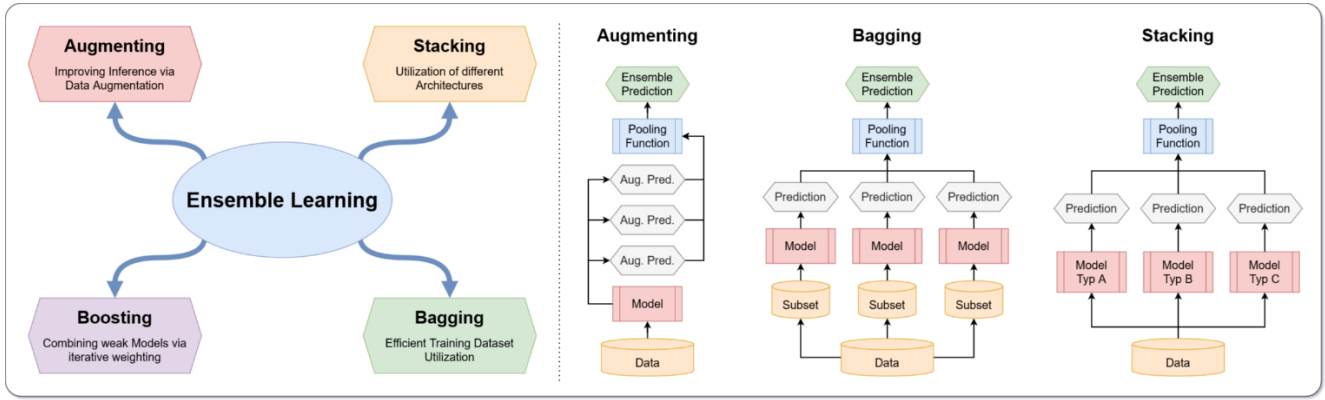

**Figure 2:** Illustration showing the four ensemble learning techniques: Augmenting, Stacking, Boosting and Bagging. Except for Boosting, all techniques are commonly utilized in modern deep convolutional neural network pipelines.

2.4.1 Augmenting: The Augmenting technique, often called test-time data augmentation, can be defined as the application of reasonable image augmentation prior to inference [56–61]. Through augmentation, multiple images of the same sample can be generated and then be used to compute multiple predictions. The aim of augmenting is to reduce the risk of incorrect predictions based on overfitting or too strict pattern learning [57,58,60]. In our analysis, we reused the Baseline models and applied random rotations as well as mirroring on all axes for inference. For each sample, 15 randomly augmented images were created, and their predictions were combined through an unweighted Mean as pooling function.

2.4.2 Stacking: In contrast to single algorithm approaches, the ensemble of different deep convolutional neural network architectures (also called inhomogeneous ensemble learning) showed strong benefits for overall performance [10,24,26,27,62]. This kind of ensemble learning is more complex and can consist of even different computer vision tasks [10,24,27]. The idea of the Stacking technique is to utilize these diverse and independent models by stacking another machine learning algorithm on top of these predictions. In our analysis, we reused the Baseline models consisting of various architectures as an ensemble for stacking the pooling functions directly on top of these inhomogeneous models.

2.4.3 Bagging: Homogeneous model ensembles can be defined as multiple models consisting of the same algorithm, hyperparameters, or architecture [17,24]. The Bagging technique is based on improved training dataset sampling and a popular homogeneous ensemble learning technique. In contrast to a standard single training/validation split, which results in a single model, Bagging consists of training multiple models on randomly drawn subsets from the dataset. In practice, a k-fold cross-validation is applied on the dataset resulting in k models [63]. In our analysis, we applied a 5-fold cross-validation for Bagging as described in the sub-section 2.2 Sampling and Preprocessing, which resulted in five models for each architecture. The predictions of these five models for a single sample were combined via multiple pooling functions.

## 2.5 Pooling Functions

In order to combine the ensemble of predictions into a single one, we studied several different methods and algorithms. A prediction consisted of the softmax normalized probability of each class for an unknown sample. For the Bagging and Stacking technique, the following pooling functions were analyzed: Best Model, Decision Tree, Gaussian Process classifier, Global Argmax, Logistic Regression, Majority Vote Soft and Hard, Unweighted and Weighted Mean, Naïve Bayes, Support Vector Machine, and k-Nearest Neighbors [64]. For the Augmenting technique, only the Unweighted Mean was used as pooling function.

Basic pooling functions were custom implemented, whereas more complex algorithms were integrated from scikit-learn [64]. The Best Model is selecting the best scoring model according to the F1 score on the 'ensemble-train' sampling set. Decision Trees were trained with Gini impurity as information gain function [65]. Gaussian Process classifier was based on Laplace approximation with a 'one-vs-rest' multi-class strategy. Global Argmax was defined as selecting the class with the highest probability across all predictions and zeroing the remaining classes. For Logistic Regression training, the 'newton-cg' solver and L2 regularization were used with a multinomial multi-class strategy [66]. The Majority Vote Soft variant sums up all probabilities per class and then softmax normalizes them across all classes, whereas the Majority Vote Hard variant utilizes traditional class voting in which the class with the highest probability is used for each prediction as vote. The Unweighted Mean straightforward averages the class probabilities across predictions, whereas the Weighted Mean performs a weighted averaging according to the achieved F1 score of the model on the 'ensemble-train' sampling set. The Naïve Bayes was implemented as the Complement variant described by Rennie et al. [67]. The Support Vector Machine classifier was based on the standard implementation from LIBSVM [68]. For the k-Nearest Neighbors classifier, a number of five neighbors was utilized.

## 2.6 Evaluation

For evaluation, we utilized the packages pandas [69], scikit-learn [64], and plotnine [70] for visualization. The performance scores were calculated class-wise and averaged by the unweighted mean. The following community-standard scores were used: Accuracy, F1-score, Sensitivity (also called True Positive Rate), False Positive Rate (FPR), and area under the receiver operating characteristic curve (AUC & ROC). The supplementary contains various additional metrics like Top-1/3-Error, Specificity, and others.

$$Accuracy = \frac{TP + TN}{TP + FP + TN + FN} \quad (2) \qquad F1 = \frac{2TP}{2TP + FP + FN} \quad (3)$$

$$Sensitivity = \frac{TP}{TP + FN} \quad (4) \qquad FPR = \frac{FP}{FP + TN} \quad (5)$$

All metrics are based on the confusion matrix for binary classification, where TP, FP, TN, and FN represent the true positive, false positive, true negative, and false negative rate, respectively [71]. For the AUC and ROC curve computation, classifier confidence for predictions was also utilized [72].

## 3. Results

The total training time of the complete analysis took around 1,215 hours with the following distribution per technique: Baseline 213 hours (17.5%), Augmenting 0.00 hours (0%), Stacking less than 0.09 hours (0%), and Bagging 1,002 hours (82 .5%). It has to be noted that the Augmenting and Stacking techniques were based on the Baseline models but did not require extensive additional training time. The Baseline revealed an average training time by mean across all architectures of 45 minutes for COVID, 47 minutes for CHMNIST, 302 minutes for ISIC, and 1,026 minutes for DRD, whereas the Vanilla architecture had the lowest training time on average across datasets with 246 minutes and ResNet101 the highest with 522 minutes. Further details on training times for all architectures and phases can be found in the supplementary.

All training processes for the deep learning convolutional neural network models did not require the entire 1000 epochs and instead were early stopped after an average of 51 epochs. On median the epoch distribution looked like the following: For Baseline CHMNIST 54, COVID 48 ISIC 64, and DRD 37. For Bagging CHMNIST 53, COVID 48, ISIC 68, and DRD 37. Through validation monitoring during the training, no overfitting was observed. The training and validation loss function revealed no significant distinction from each other. The individual fitting plots for all models are attached in the Appendix.

### 3.1 Baseline

The Baseline revealed the performance of various state-of-the-art architectures without the usage of any ensemble learning technique. This resulted in an average F1-score by a median of 0.95 for CHMNIST, 0.96 for COVID, 0.72 for ISIC, and 0.43 for DRD. The architectures shared overall a similar performance depending on the dataset noisiness. According to their F1-score, the best architectures were EfficientNetB4 and ResNet101 in CHMNIST, ResNeXt101 in COVID, ResNet101 and ResNeXt101 in ISIC, as well as EfficientNetB4 and ResNet101 in DRD. The smaller architectures like Vanilla and MobileNetV2 performed the worst. More details are shown in Table 2.

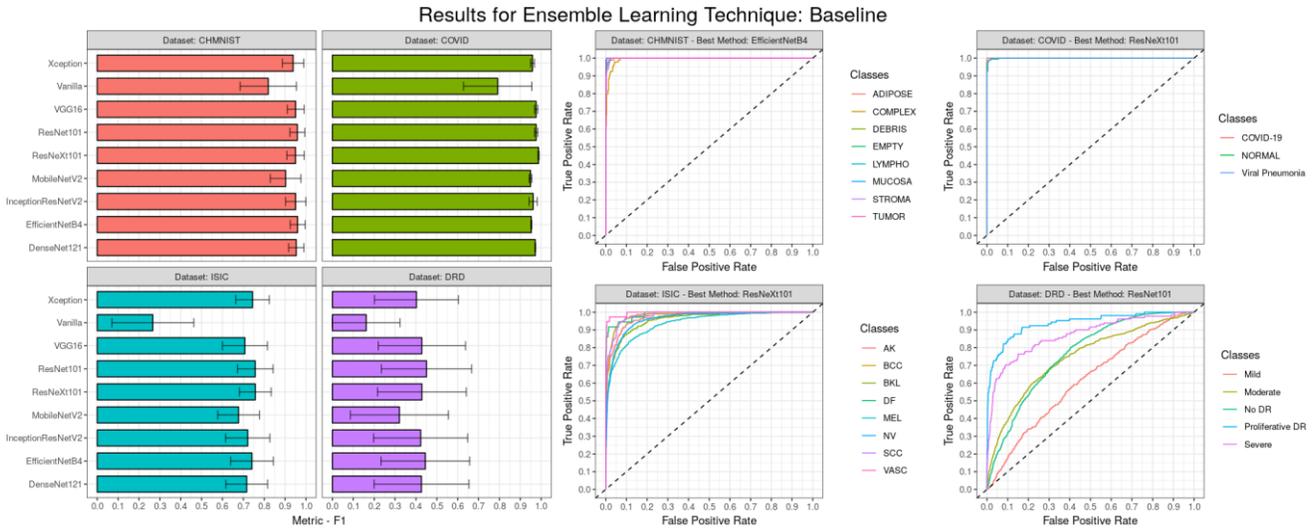

**Figure 3:** Performance results for the Baseline pipeline showing (LEFT) the achieved F1-score with its confidence interval on classification for each architecture and (RIGHT) class-wise ROC curves for the best performing architecture (according to F1-score) on each dataset.

The receiver operating characteristic curves (ROC) curves in Figure 3 revealed only marginal performance differences of classes in the CHMNIST, COVID, and ISIC dataset. However, DRD showed significant differences in Accuracy between classes whereas the detection of 'Mild' samples had the lowest performance.

**Table 2:** Achieved results of the Baseline approach showing the Accuracy (Acc.), F1-score, Sensitivity (Sens.), and AUC on image classification for each architecture and dataset.

| | CHMNIST | | | | COVID | | | | ISIC | | | | DRD | | | |
|---|---|---|---|---|---|---|---|---|---|---|---|---|---|---|---|---|
| **Method** | **Acc.** | **F1** | **Sens.** | **AUC** | **Acc.** | **F1** | **Sens.** | **AUC** | **Acc.** | **F1** | **Sens.** | **AUC** | **Acc.** | **F1** | **Sens.** | **AUC** |
| **DenseNet121** | 0.99 | 0.95 | 0.95 | 1.0 | 0.98 | 0.97 | 0.97 | 1.0 | 0.95 | 0.72 | 0.76 | 0.96 | 0.84 | 0.43 | 0.48 | 0.78 |
| **EfficientNetB4** | 0.99 | **0.96** | 0.96 | 1.0 | 0.97 | 0.95 | 0.96 | 0.99 | 0.95 | 0.74 | 0.79 | 0.97 | 0.84 | **0.45** | 0.53 | 0.81 |
| **Inception-Res-NetV2** | 0.99 | 0.95 | 0.95 | 1.0 | 0.98 | 0.96 | 0.96 | 1.0 | 0.95 | 0.72 | 0.74 | 0.96 | 0.84 | 0.42 | 0.5 | 0.78 |
| **MobileNetV2** | 0.98 | 0.9 | 0.9 | 0.99 | 0.97 | 0.95 | 0.96 | 0.99 | 0.94 | 0.68 | 0.73 | 0.95 | 0.82 | 0.32 | 0.41 | 0.72 |
| **ResNeXt101** | 0.99 | 0.95 | 0.95 | 1.0 | 0.99 | **0.99** | 0.98 | 1.0 | 0.96 | **0.76** | 0.78 | 0.97 | 0.83 | 0.43 | 0.52 | 0.79 |
| **ResNet101** | 0.99 | **0.96** | 0.96 | 1.0 | 0.98 | 0.98 | 0.98 | 1.0 | 0.95 | **0.76** | 0.79 | 0.97 | 0.84 | **0.45** | 0.48 | 0.79 |
| **VGG16** | 0.99 | 0.95 | 0.95 | 1.0 | 0.98 | 0.98 | 0.97 | 1.0 | 0.94 | 0.71 | 0.74 | 0.96 | 0.82 | 0.43 | 0.49 | 0.78 |
| **Vanilla** | 0.95 | 0.82 | 0.82 | 0.98 | 0.91 | 0.79 | 0.79 | 0.94 | 0.86 | 0.27 | 0.34 | 0.8 | 0.74 | 0.16 | 0.27 | 0.59 |
| **Xception** | 0.98 | 0.94 | 0.94 | 1.0 | 0.97 | 0.96 | 0.95 | 1.0 | 0.95 | 0.74 | 0.77 | 0.96 | 0.82 | 0.4 | 0.48 | 0.78 |

### 3.2 Augmenting

By integrating the ensemble learning technique Augmenting for the inference based on the Baseline models, it was possible to obtain the following average F1-scores by median: 0.95 for CHMNIST, 0.97 for COVID, 0.74 for ISIC, and 0.43 for DRD. More details are shown in Table 3. Thus, there was only a marginal performance increase for the CHMNIST and ISIC dataset compared to the Baseline. However, in the comparison of the best possible score between Augmenting and Baseline, a performance impact of 0% for CHMNIST, -1% for COVID, +3% for ISIC, and +4% for DRD was measured according to the F1-score.

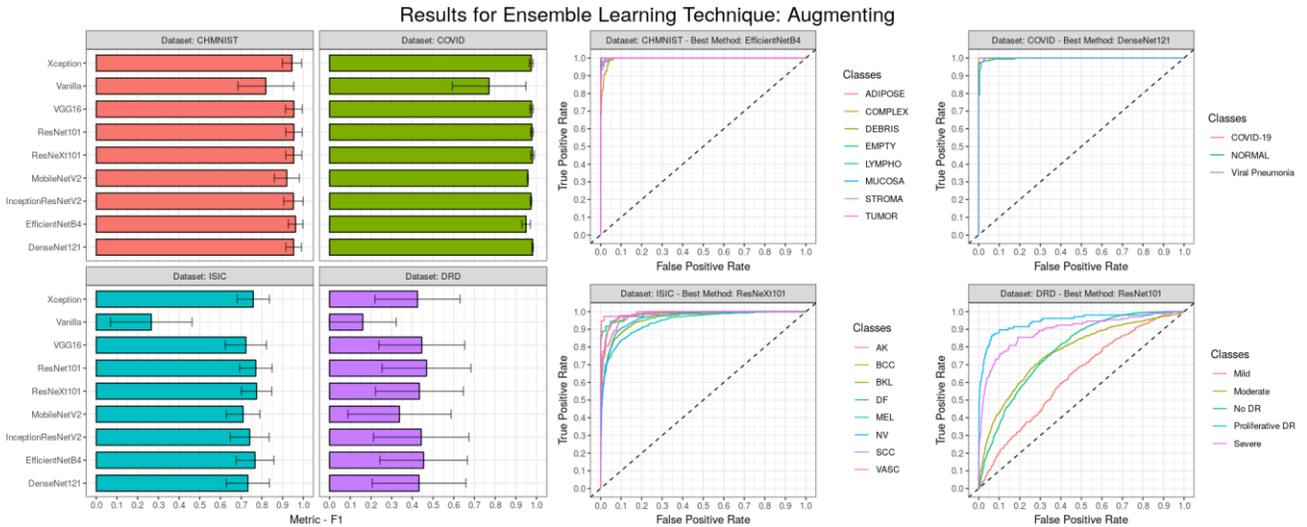

**Figure 4:** Performance results for the Augmenting approach showing (LEFT) the achieved F1-score with its confidence interval on classification for each architecture and (RIGHT) class-wise ROC curves for the best performing architecture (according to F1-score) on each dataset.

The ranking between best-performing architectures revealed no drastic change. Especially, the EfficientNetB4 and ResNet101 achieved the highest performance similar to the Baseline, as well as the smaller architectures like Vanilla and MobileNetV2 the lowest. The ROC curves in Figure 4 resulted in equivalent model Accuracy variance between classes and datasets as the Baseline.

**Table 3:** Achieved results of the Augmenting approach showing the Accuracy (Acc.), F1-score, Sensitivity (Sens.), and AUC on image classification for each architecture and dataset.

| Method | CHMNIST | | | | COVID | | | | ISIC | | | | DRD | | | |
|---|---|---|---|---|---|---|---|---|---|---|---|---|---|---|---|---|
| | Acc. | F1 | Sens. | AUC | Acc. | F1 | Sens. | AUC | Acc. | F1 | Sens. | AUC | Acc. | F1 | Sens. | AUC |
| **DenseNet121** | 0.99 | 0.95 | 0.95 | 1.0 | 0.99 | **0.98** | 0.98 | 1.0 | 0.95 | 0.73 | 0.77 | 0.97 | 0.85 | 0.43 | 0.5 | 0.79 |
| **EfficientNetB4** | 0.99 | **0.96** | 0.96 | 1.0 | 0.97 | 0.95 | 0.95 | 0.99 | 0.96 | **0.77** | 0.81 | 0.97 | 0.85 | 0.45 | 0.55 | 0.82 |
| **Inception-Res-NetV2** | 0.99 | 0.95 | 0.95 | 1.0 | 0.98 | 0.97 | 0.97 | 1.0 | 0.95 | 0.74 | 0.76 | 0.97 | 0.85 | 0.44 | 0.52 | 0.79 |
| **MobileNetV2** | 0.98 | 0.92 | 0.92 | 0.99 | 0.97 | 0.96 | 0.96 | 0.99 | 0.94 | 0.71 | 0.76 | 0.96 | 0.84 | 0.34 | 0.42 | 0.74 |
| **ResNeXt101** | 0.99 | 0.95 | 0.95 | 1.0 | 0.99 | **0.98** | 0.98 | 1.0 | 0.96 | **0.77** | 0.8 | 0.97 | 0.84 | 0.43 | 0.52 | 0.8 |
| **ResNet101** | 0.99 | 0.95 | 0.95 | 1.0 | 0.98 | **0.98** | 0.98 | 1.0 | 0.95 | **0.77** | 0.8 | 0.97 | 0.85 | **0.47** | 0.51 | 0.8 |
| **VGG16** | 0.99 | 0.95 | 0.95 | 1.0 | 0.98 | **0.98** | 0.97 | 1.0 | 0.95 | 0.72 | 0.75 | 0.96 | 0.83 | 0.45 | 0.51 | 0.79 |
| **Vanilla** | 0.95 | 0.82 | 0.82 | 0.98 | 0.9 | 0.77 | 0.8 | 0.93 | 0.86 | 0.27 | 0.35 | 0.8 | 0.74 | 0.16 | 0.26 | 0.59 |
| **Xception** | 0.99 | 0.95 | 0.95 | 1.0 | 0.98 | 0.97 | 0.97 | 1.0 | 0.95 | 0.76 | 0.77 | 0.97 | 0.83 | 0.42 | 0.5 | 0.8 |

### 3.3 Stacking

For the Stacking technique, several pooling functions were successfully applied for combining the predictions of all Baseline architectures and resulted in the following average F1-scores by median: 0.96 for CHMNIST, 0.98 for COVID, 0.81 for ISIC, and 0.48 for DRD. More details are shown in Table 4. Compared with the median F1-score of the Baseline, a performance impact of +1% for CHMNIST, +2% for COVID, +13% for ISIC, and +12% for DRD was measured. Additional to the median performance comparison, the pooling function 'Best Model' was also used as a benchmark without the usage of ensemble learning, which was inferior of up to 0.08 in Accuracy, 0.06 in F1, 0.06 in Sensitivity, and 0.04 in AUC compared with the best pooling function.

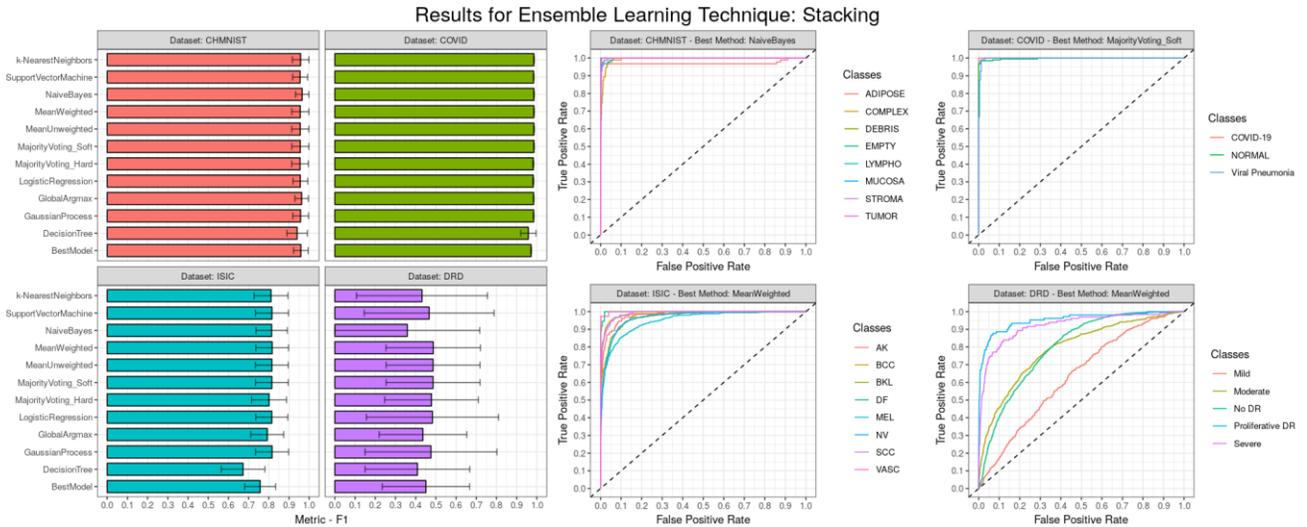

**Figure 5**: Performance results for the Stacking approach showing (LEFT) the achieved F1-score with its confidence interval on classification for each pooling function and (RIGHT) class-wise ROC curves for the best performing method (according to F1-score) on each dataset.

According to their F1-score, the best pooling functions were Naïve Bayes in CHMNIST, Majority Voting Soft, Mean Un-/Weighted, Naïve Byes and k-Nearest Neighbor in COVID, Gaussian Process, Majority Voting Soft, Mean Un-/Weighted and Support Vector Machine in ISIC, as well as Majority Voting Soft and Mean Un-/Weighted in DRD. The Decision Tree pooling function performed the worst and had a performance difference of up to -0.02 Accuracy, -0.09 F1, -0.09 Sensitivity, and -0.15 AUC compared to the 'Best Model' from the Baseline.

The ROC curves of the Stacking approach (illustrated in Figure 5) showed the same trend of class-wise performance differences as the Baseline, but with better precision results especially in the ISIC and DRD dataset.

**Table 4:** Achieved results of the Stacking approach showing the Accuracy (Acc.), F1-score, Sensitivity (Sens.), and AUC on image classification for each technique and dataset.

| Method | CHMNIST | | | | COVID | | | | ISIC | | | | DRD | | | |
|---|---|---|---|---|---|---|---|---|---|---|---|---|---|---|---|---|
| | Acc. | F1 | Sens. | AUC | Acc. | F1 | Sens. | AUC | Acc. | F1 | Sens. | AUC | Acc. | F1 | Sens. | AUC |
| **Best Model** | 0.99 | 0.96 | 0.96 | 1.0 | 0.98 | 0.97 | 0.97 | 1.0 | 0.96 | 0.76 | 0.78 | 0.97 | 0.84 | 0.45 | 0.48 | 0.79 |
| **Decision Tree** | 0.98 | 0.94 | 0.94 | 0.97 | 0.98 | 0.96 | 0.97 | 0.98 | 0.94 | 0.67 | 0.69 | 0.82 | 0.87 | 0.41 | 0.42 | 0.64 |
| **Gaussian Process** | 0.99 | 0.96 | 0.96 | 0.99 | 0.99 | 0.98 | 0.98 | 1.0 | 0.96 | **0.82** | 0.79 | 0.97 | 0.92 | 0.48 | 0.44 | 0.83 |
| **Global Argmax** | 0.99 | 0.96 | 0.96 | 1.0 | 0.99 | 0.98 | 0.98 | 1.0 | 0.96 | 0.79 | 0.82 | 0.94 | 0.85 | 0.44 | 0.54 | 0.71 |
| **Logistic Regression** | 0.99 | 0.96 | 0.96 | 1.0 | 0.99 | 0.98 | 0.98 | 1.0 | 0.96 | 0.81 | 0.79 | 0.98 | 0.92 | 0.48 | 0.45 | 0.83 |
| **Majority Voting Hard** | 0.99 | 0.95 | 0.95 | 0.99 | 0.99 | 0.98 | 0.98 | 1.0 | 0.96 | 0.8 | 0.81 | 0.96 | 0.87 | 0.48 | 0.52 | 0.79 |
| **Majority Voting Soft** | 0.99 | 0.96 | 0.96 | 1.0 | 0.99 | **0.99** | 0.98 | 1.0 | 0.96 | **0.82** | 0.82 | 0.97 | 0.87 | **0.49** | 0.53 | 0.81 |
| **Mean Unweighted** | 0.99 | 0.96 | 0.96 | 1.0 | 0.99 | **0.99** | 0.98 | 1.0 | 0.96 | **0.82** | 0.82 | 0.98 | 0.87 | **0.49** | 0.53 | 0.82 |
| **Mean Weighted** | 0.99 | 0.96 | 0.96 | 1.0 | 0.99 | **0.99** | 0.98 | 1.0 | 0.96 | **0.82** | 0.82 | 0.98 | 0.87 | **0.49** | 0.53 | 0.82 |
| **Naive Bayes** | 0.99 | **0.97** | 0.97 | 0.99 | 0.99 | **0.99** | 0.98 | 1.0 | 0.96 | 0.81 | 0.81 | 0.97 | 0.9 | 0.36 | 0.41 | 0.79 |
| **Support Vector Machine** | 0.99 | 0.96 | 0.95 | 1.0 | 0.99 | 0.98 | 0.98 | 1.0 | 0.96 | **0.82** | 0.8 | 0.97 | 0.92 | 0.47 | 0.42 | 0.8 |
| **k-Nearest Neighbor** | 0.99 | 0.96 | 0.96 | 0.99 | 0.99 | **0.99** | 0.98 | 0.99 | 0.96 | 0.81 | 0.79 | 0.93 | 0.91 | 0.43 | 0.4 | 0.74 |

### 3.4 Bagging

By training new models based on a 5-fold cross-validation, it was possible to analyze the effects of Bagging on prediction capability. The predictions of five models per architecture were combined using various pooling functions. In this experiment, the five models of the EfficientNetB4 architecture archived the highest F1-scoring and were selected for further result reporting and representation of the Bagging approach. The evaluation of the merged predictions of these models showed the following averaged F1-score results by median: 0.96 for CHMNIST, 0.98 for COVID, 0.8 for ISIC, and 0.47 for DRD. In comparison with the Baseline, the following performance impact was measured: +1% for CHMNIST, +2% for COVID, +11% for ISIC, and +9% for DRD. More details for the Bagging results can be seen in Table 5.

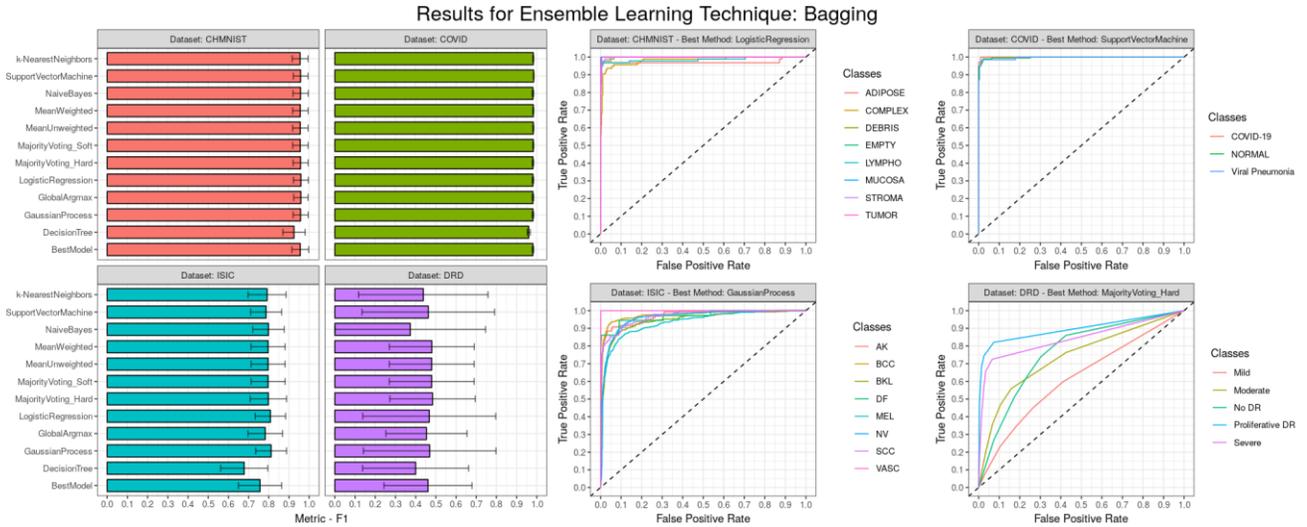

**Figure 6:** Performance results for the Bagging approach showing (LEFT) the achieved F1-score with its confidence interval on classification for each pooling function and (RIGHT) class-wise ROC curves for the best performing method (according to F1-score) on each dataset. The Bagging technique was applied on the EfficientNetB4 architecture, which showed the highest F1-score performance across all architectures.

On the contrary to the previous ensemble learning approaches, the 'Best Model' pooling function represents not the best validation scoring Baseline model but instead the best model from the 5-fold cross-validation. The ranking between best-performing pooling functions for the EfficientNetB4 5-fold cross-validation revealed close grouping around the same score. In the CHMNIST and COVID set, all pooling functions except for Decision Trees achieved an F1-score of 0.96 and 0.98, respectively. Overall, the pooling based on Mean, Majority Voting, Gaussian Process, and Logistic Regression resulted in the highest performance on average. On the other hand, Decision Tree and Naïve Bayes obtained the lowest F1-scores.

In Figure 6, the ROC curves showed inferior performance compared to the Baseline. Notably, the CHMNIST and COVID curves reveal a strong precision decrease. Furthermore, the ISIC dataset indicates a stronger model Accuracy variance between classes compared to the Baseline ROC curves.

**Table 5:** Achieved results of the Bagging approach showing the Accuracy (Acc.), F1-score, Sensitivity (Sens.), and AUC on image classification for each technique and dataset. The Bagging technique was applied on the EfficientNetB4 architecture, which showed the highest F1-score performance.

| | CHMNIST | | | | COVID | | | | ISIC | | | | DRD | | | |
|---|---|---|---|---|---|---|---|---|---|---|---|---|---|---|---|---|
| **Method** | Acc. | F1 | Sens. | AUC | Acc. | F1 | Sens. | AUC | Acc. | F1 | Sens. | AUC | Acc. | F1 | Sens. | AUC |
| **Best Model** | 0.99 | **0.96** | 0.96 | 1.0 | 0.99 | **0.98** | 0.98 | 1.0 | 0.95 | 0.76 | 0.78 | 0.97 | 0.86 | 0.46 | 0.53 | 0.82 |
| **Decision Tree** | 0.98 | 0.92 | 0.93 | 0.96 | 0.97 | 0.96 | 0.95 | 0.96 | 0.94 | 0.68 | 0.66 | 0.81 | 0.86 | 0.4 | 0.41 | 0.63 |
| **Gaussian Process** | 0.99 | **0.96** | 0.96 | 0.99 | 0.99 | **0.98** | 0.98 | 1.0 | 0.96 | **0.81** | 0.8 | 0.97 | 0.92 | 0.47 | 0.46 | 0.83 |
| **Global Argmax** | 0.99 | **0.96** | 0.96 | 0.99 | 0.99 | **0.98** | 0.98 | 1.0 | 0.96 | 0.78 | 0.82 | 0.95 | 0.84 | 0.45 | 0.56 | 0.72 |
| **Logistic Regression** | 0.99 | **0.96** | 0.96 | 0.99 | 0.99 | **0.98** | 0.98 | 1.0 | 0.96 | **0.81** | 0.8 | 0.97 | 0.92 | 0.47 | 0.46 | 0.84 |
| **Majority Voting Hard** | 0.99 | **0.96** | 0.96 | 0.98 | 0.99 | **0.98** | 0.98 | 0.99 | 0.96 | 0.8 | 0.81 | 0.94 | 0.85 | **0.48** | 0.55 | 0.77 |
| **Majority Voting Soft** | 0.99 | **0.96** | 0.96 | 1.0 | 0.99 | **0.98** | 0.98 | 1.0 | 0.96 | 0.8 | 0.82 | 0.97 | 0.85 | **0.48** | 0.56 | 0.81 |
| **Mean Unweighted** | 0.99 | **0.96** | 0.96 | 1.0 | 0.99 | **0.98** | 0.98 | 1.0 | 0.96 | 0.8 | 0.82 | 0.98 | 0.85 | **0.48** | 0.56 | 0.83 |
| **Mean Weighted** | 0.99 | **0.96** | 0.96 | 1.0 | 0.99 | **0.98** | 0.98 | 1.0 | 0.96 | 0.8 | 0.82 | 0.98 | 0.85 | **0.48** | 0.56 | 0.83 |
| **Naive Bayes** | 0.99 | **0.96** | 0.96 | 0.99 | 0.99 | **0.98** | 0.98 | 1.0 | 0.96 | 0.8 | 0.82 | 0.97 | 0.9 | 0.37 | 0.43 | 0.8 |
| **Support Vector Machine** | 0.99 | **0.96** | 0.96 | 0.99 | 0.99 | **0.98** | 0.98 | 1.0 | 0.96 | 0.79 | 0.76 | 0.95 | 0.92 | 0.46 | 0.43 | 0.79 |
| **k-Nearest Neighbor** | 0.99 | **0.96** | 0.95 | 0.99 | 0.99 | **0.98** | 0.98 | 0.99 | 0.96 | 0.79 | 0.78 | 0.92 | 0.91 | 0.44 | 0.43 | 0.74 |

## 4. Discussion

In this work, we setup a reproducible pipeline for analyzing the impact of ensemble learning techniques on MIC performance with deep convolutional neural networks. We implemented Augmenting, Bagging as well as Stacking and compared them to a Baseline to compute performance gain on various metrics like F1-score, Sensitivity, AUC, and Accuracy. Our analysis proved that the integration of ensemble learning techniques can significantly boost classification performance from deep convolutional neural network models. As in Figure 7 summarized, our results showed a performance gain ranking from highest to lowest for the following ensemble learning techniques: Stacking, Bagging, and Augmenting.

  The ensemble learning technique with the highest performance gain was Stacking, which applies pooling functions on top of different deep convolutional neural network architectures. Various state-of-the-art MIC pipelines heavily utilize a Stacking based pipeline structure to optimize performance by combining novel architectures or differently trained models [6,10,13,21,73]. This results in higher inference quality and bias or error reduction by using the prediction information of diverse methods. Our analysis also revealed that, according to F1-score results, simple pooling functions like averaging by Mean or a Soft Majority Vote results in an equally strong or even higher performance gain compared to more complex pooling functions like Support Vector Machines or Logistic Regressions. However, according to Accuracy results, the more complex pooling functions obtained higher scores. This indicates that more simple pooling functions are still based on the penalty strategy of the models which were trained with a class weighted loss function in our experiments. Thus, our results of simple pooling functions still optimize for class balanced metrics like F1-score or Sensitivity. On the other hand, more complex pooling functions with a separated training process focused on optimizing overall true cases including true negatives which resulted in better scoring on unbalanced metrics like Accuracy. Apart from that, other recent studies which analyzed the impact of Stacking also support our hypothesis that Stacking can significantly improve individual deep convolutional neural network model performance by up to 10% [7,14,24,27]. With a similar experiment design as in our work, Kandel et al. demonstrated Stacking impact on a musculoskeletal fracture dataset analyzing pooling functions based on statistics as well as probability [27].

  The Augmenting technique demonstrated to be an efficient ensemble learning approach. In nearly all our experiments, it was possible to improve the performance by another few percent through reducing overfitting bias in predictions. In theory, this should be already avoided with standard data augmentation during the training

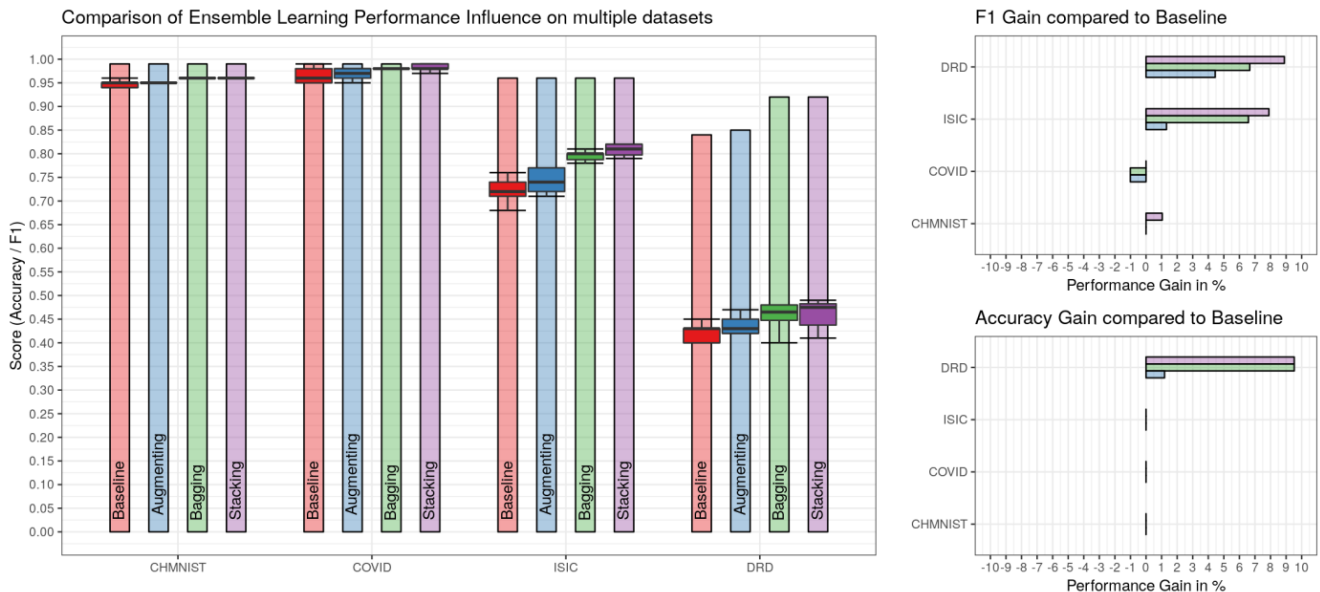

**Figure 7:** Summary of all experiments to identify performance impact of ensemble learning techniques on medical image classification.

LEFT: Bar plots showing the maximum achieved Accuracy across all methods for each ensemble learning technique and dataset: Baseline (red), Augmenting (blue), Bagging (green) and Stacking (purple). Additionally, the distribution of achieved F1-scores by the various methods is illustrated with box plots.

RIGHT: Computed performance impact between the best scoring method of the Baseline and the best scoring method of the applied ensemble learning technique for each dataset. The performance impact is represented as performance gain in % between F1-scores (RIGHT TOP) as well as Accuracies (RIGHT BOTTOM). The color mapping of the ensemble learning techniques are equal to Figure 7 LEFT (Augmenting: Blue; Bagging: Green; Stacking: Purple).

process. Although, our experiments indicated that the increased image variability through Augmenting could lead to adverse performance influences if applied on models based on small-sized datasets with a high risk of being overfitted. Especially in medical imaging, in which small datasets are common, this effect should be considered if Augmenting is applied and can also act as a strong indicator for overfitting. Nevertheless, strong performing MIC pipelines revealed that model performance can be significantly boosted with inference Augmenting [58–61]. Recent studies from Kandel et al. [60] and Shanmugam et al. [58] also analyzed the performance impact in detail of Augmenting on MIC and proved strong as well as consistent improvement, especially for low scoring models. In contrast to other ensemble learning techniques, Augmenting can be quickly integrated into pipelines without the need for additional training of various deep convolutional neural network or machine learning models. Thus, also a single model pipeline can benefit from this ensemble learning technique. However, performance gain from Augmenting is strongly influenced by applied augmentation methods and medical context in a dataset. Molchanov et al. tried to solve this issue with a greedy policy search to find the optimal Augmentation configuration [61].

Nowadays, Bagging is one of the most widely used ensemble learning techniques and utilized in several state-of-the-art pipelines and top-performing benchmark submissions in MIC [12,14,15,21,24,74]. In compliance, our experiments Bagging showed a strong performance increase for large datasets and no or marginal performance decrease in small datasets. Similar to Stacking, Bagging was able to significantly improve prediction capability for complex datasets like ISIC and DRD. We interpreted the possible detrimental effects in COVID and CHMNIST that the fewer data used for model training through cross-validation sampling had a considerable impact on performance in smaller datasets. Especially in small medical datasets with rare and unique morphological cases, excluding these can have a strong negative impact on performance. This is why our large datasets like ISIC and DRD with adequate feature presentations in all sampled folds revealed persistent performance improvement. Studies like Dwork et al. [75] analyzed this behavior and concluded that cross-validation based strategies comprise sustainable overfitting risk [76]. Based on our results, Bagging showed to have a high risk of drifting away from an optimal bias-variance tradeoff. According to Geman et al. [77], the bias-variance tradeoff is the right balance between bias and variance in a machine learning model in order to obtain the optimal generalizable model [77]. Whereas increased bias results into the risk of underfitting, increased variance can lead to overfitting. Cross-validation based Bagging boosts efficient data usage and, thus, the variance of a model. However, it has to be noted that the bias-variance tradeoff is still on active discussion in the research community for its correctness in deep learning [78,79]. Furthermore, Bagging requires extensive additional training time to obtain multiple models. In the field of deep learning, training a higher number of models can lead to an extremely time-consuming process. For this reason, we specified our analysis on a 5-fold cross-validation. Still, further research is needed on the impact of fold number or sampling size on performance and model generalizability in deep learning based MIC. Nevertheless, we concluded that Bagging is a powerful but complex to utilize ensemble learning technique and that its effectiveness is highly depended on sufficient feature representation in the sampled cross-validation folds. To avoid harmful folds with missing feature representation, we promote in-detail dataset analysis with manual annotation supported sampling (stratified) or using a higher k-fold to increase training sets and, thus, reduce the risk of excluding samples with unique morphological features.

## 5. Conclusions

In this paper, we analyzed the impact of the most widely used ensemble learning techniques on medical image classification performance: Augmenting, Stacking, and Bagging. We setup a reproducible experiment pipeline, evaluated the performance through multiple metrics, and compared these techniques with a Baseline to identify possible performance gain. Our results revealed that Stacking was able to achieve the largest performance gain in our medical image classification pipeline. Augmenting showed consistent improvement capabilities on non-overfitting models and has the advantage to be applicable to also single model based pipelines. Cross-validation based Bagging demonstrated significant performance gain close to Stacking, but reliant on sampling with sufficient feature representation in all folds. Additionally, we showed that simple statistical pooling functions like Mean or Majority Voting are equal or often even better than more complex pooling functions like Support Vector Machines. Overall, we concluded that the integration of ensemble learning techniques is a powerful method for MIC pipeline improvement and performance boosting. As future research, we plan to further analyze the impact of the number of folds in cross-validation based Bagging techniques and extend our analysis on deep learning Boosting approaches. Furthermore, the applicability of explainable artificial intelligence techniques for ensemble learning based medical image classification pipelines with multiple models is still an open research field and requires further research.

### Acknowledgments
We want to thank Edmund Müller, Dennis Hartmann, Philip Meyer, and Peter Parys for their useful comments and support. We also want to thank Johannes Raffler and Ralf Huss for sharing their GPU hardware with us which was used for this work.

### Conflicts of Interest
None declared.

### Funding
This work is a part of the DIFUTURE project funded by the German Ministry of Education and Research (Bundesministerium für Bildung und Forschung, BMBF) grant FKZ01ZZ1804E.

### Appendix
The code for this article was implemented in Python (platform independent) and is available under the GPL-3.0 License at the following GitHub repository: https://github.com/frankkramer-lab/ensmic.

All data generated and analyzed during this study is available in the following Zenodo repository: https://doi.org/10.5281/zenodo.6457912.